\documentclass{article}

\PassOptionsToPackage{numbers,compress}{natbib}

 \usepackage[dblblindworkshop, final]{neurips_2025}

\usepackage[utf8]{inputenc} 
\usepackage[T1]{fontenc}    
\usepackage{hyperref}       
\usepackage{url}            
\usepackage{booktabs}       
\usepackage{amsfonts}   
\usepackage{algorithm}
\usepackage{algpseudocode}
\usepackage{nicefrac}       
\usepackage{microtype}      
\usepackage{xcolor}         
\usepackage{amsmath}
\usepackage{graphicx}
\usepackage{subcaption}
\usepackage{booktabs}
\usepackage{subcaption}
\usepackage[export]{adjustbox}   
\usepackage{graphicx}
\usepackage{subcaption}
\usepackage[labelfont=bf]{caption}
\usepackage{amsmath,amssymb,amsthm}

\newtheorem{definition}{Definition}

\newcommand{\F}{\boldsymbol{F}}
\newcommand{\g}{\boldsymbol{g}}
\newcommand{\boldu}{\boldsymbol{u}}
\newcommand{\x}{\boldsymbol{x}}
\newcommand{\w}{\boldsymbol{w}}
\newcommand{\boldv}{\boldsymbol{v}}

\newcommand{\T}{\mathcal{T}}
\newcommand{\D}{\mathcal{D}}

\usepackage{amsthm}

\newtheorem{theorem}{Theorem}[section]

\newtheorem{lemma}[theorem]{Lemma}

\title{ONG: Orthogonal Natural Gradient Descent}

\author{
Yajat Yadav \\
UC Berkeley \\
\texttt{yajatyadav@berkeley.edu}
\And
Patrick Mendoza \\
UC Berkeley \\
\texttt{patmendoza6745@berkeley.edu}
\And
Jathin Korrapati \\
UC Berkeley \\
\texttt{jkorr@berkeley.edu}
}

\begin{document}

\maketitle
\begin{abstract}
    Orthogonal Gradient Descent (OGD) has emerged as a powerful method for continual learning. However, its Euclidean projections do not leverage the underlying information-geometric structure of the problem, which can lead to suboptimal convergence in learning tasks. To address this, we propose incorporating the natural gradient into OGD and present \textbf{ONG (Orthogonal Natural Gradient Descent)}. ONG preconditions each new task-specific gradient with an efficient EKFAC approximation of the inverse Fisher information matrix, yielding updates that follow the steepest descent direction under a Riemannian metric. To preserve performance on previously learned tasks, ONG projects these natural gradients onto the orthogonal complement of prior tasks' natural gradients. We provide an initial theoretical justification for this procedure, introduce the Orthogonal Natural Gradient Descent (ONG) algorithm, and present preliminary results on the Permuted and Rotated MNIST benchmarks. Our preliminary results, however, indicate that a naive combination of natural gradients and orthogonal projections has potential issues. This finding has motivated continued future work focused on robustly reconciling these geometric perspectives to develop a continual learning method, establishing a more rigorous theoretical foundation with formal convergence guarantees, and extending empirical validation to large-scale continual learning benchmarks. The anonymized version of our code can be found as the zip file \href{https://drive.google.com/drive/folders/11PyU6M8pNgOUB5pwdGORtbnMtD8Shiw_?usp=sharing}{here}.
    
\end{abstract}

\section{Introduction}
Continual learning, the process of training a single model on a sequence of tasks without forgetting previously learned tasks, is a major challenge in deep learning. Naive fine-tuning leads to catastrophic forgetting, where earlier tasks' performance collapses as networks shift their weights to accommodate the new task, forgetting previously learned tasks  \cite{MCCLOSKEY1989109, goodfellow2015empiricalinvestigationcatastrophicforgetting}. In deep learning, a promising training-time algorithm to mitigate this issue is Orthogonal Gradient Descent (OGD) \cite{OGD}. OGD works by projecting each task gradient to an orthogonal subspace spanned by previous tasks' gradients. By subtracting off this projection before taking the gradient step, we ensure that the model retains performance on previously seen tasks. Another idea in optimization is the natural gradient. Natural gradients represent the steepest descent direction with respect to the underlying geometry of the parameter space. Iteratively following the natural gradient yields the natural gradient descent algorithm \cite{amari1998natural}.
In many applications, Natural Gradient Descent requires far fewer iterations to converge than standard gradient descent. It is also invariant to any smooth and invertible reparameterization of the model, as compared to gradient descent, which is parametrization dependent \cite{martens2020new}.
 Natural Gradient Descent works by modifying the gradient update rule to account for the "information geometry" of the parameter space. In the context of probabilistic models, this involves preconditioning the gradient in the update rule with the inverse Fisher information matrix (since the Riemannian metric structure of the parameter space is described by the Fisher information \cite{amari1998natural}). Natural Gradient Descent methods have been used in several applications, such as reinforcement learning \cite{NIPS2001_4b86abe4} and variational inference \cite{wu2024understandingstochasticnaturalgradient}.

In this work, we present \textbf{Orthogonal Natural Gradient Descent (ONG)}, a novel algorithm for continual learning that integrates natural gradients within the Orthogonal Gradient Descent (OGD) framework. ONG is designed to harness the complementary strengths of these methods: the explicit forgetting prevention of OGD and the efficient, geometrically-informed updates of natural gradient optimization. We formalize the ONG algorithm, establish preliminary theoretical guarantees for its performance, and provide an efficient implementation. However, our empirical results on standard continual learning benchmarks suggests that naively combining natural gradients with orthogonal projections not only doesn't mitigate catastrophic forgetting, but also seems to worsen the performance of vanilla OGD. This suggests some fundamental geometric incompatibility between Euclidean projections and the Fisher-metric space, and has inspired us to continue future work on rigorously developing and justifying the correct way to develop geometry-consistent projection operators. We are currently working on investigating the exact failure modes of our method to better understand why this naive combination of two ideas starts to degrade performance. In addition, we are simultaneously working on extending our evaluation criteria to incorporate more challenging and realistic continual learning settings. 

\section{Related Work}
\label{sec:related}

\paragraph{Continual Learning Methods} Continual learning methods broadly fall into three categories: regularization-based, replay-based, and dynamic architecture methods. Regularization-based methods add task-specific penalties to the loss to preserve old knowledge. Notable works include Elastic Weight Consolidation \cite{Kirkpatrick_2017}, which uses the Fisher information at a task's optimum as a quadratic penalty on parameter changes, and Synaptic Intelligence (SI) \cite{zenke2017continuallearningsynapticintelligence}, which integrates a Hebbian-like importance of each weight over time. Replay-based methods mitigate forgetting by rehearsing data from previous tasks during new-task training. This can be done by either storing real data or generating synthetic examples that approximate prior distributions. For example, deep Generative Replay (DGR) methods \cite{shin2017continuallearningdeepgenerative} train a generative model to mimic old data distributions. During new task training, the generative model samples pseudo-data from prior tasks, allowing the learner to ``replay'' previous experiences without storing real examples. Finally, dynamic architecture methods seek to address catastrophic forgetting by modifying the model structure over time, often by allocating task-specific sub-modules or expanding capacity to accommodate new tasks. For instance, Progressive Neural Networks (PNNs) \cite{rusu2022progressiveneuralnetworks} allocate a new column for each task and connect it to previous columns via lateral connections. This guarantees no forgetting but scales poorly with the number of tasks. 

A notable continual learning method, which inspires the foundation of our work, is Orthogonal Gradient Descent (OGD) \cite{OGD}. OGD mitigates forgetting by projecting incoming task gradients onto the orthogonal complement of the subspace spanned by previous tasks' observed gradients. Follow-up work by Bennani et al. \cite{bennani} offers theoretical guarantees for OGD in the neural tangent kernel (NTK) regime, demonstrating that under mild assumptions, OGD can preserve task performance indefinitely.

\paragraph{Natural Gradient Methods}
 Natural Gradient Descent \cite{amari1998natural} generalizes vanilla gradient descent by taking steps that respect the geometry of the parameter space. Instead of updating in the Euclidean parameter-space, natural gradients perform optimization in the space of probability distributions, where the Fisher information matrix (FIM) serves as a Riemannian metric. In probabilistic models and Bayesian neural networks, natural gradient descent yields faster convergence and more stable updates by accounting for parameter sensitivity \cite{khan2017conjugatecomputationvariationalinference, martens2020new}. 
 



\section{Preliminaries}

\subsection{Continual Learning}

In continual learning, a model encounters tasks \(\T_1, \T_2, \T_3,...\) sequentially, each of which is its own supervised learning problem. Once the data for task \(\T_{k}\) has been utilized for training, it is usually discarded (each example is seen once). The challenge lies in updating the model so that it performs well on the current task without sacrificing its ability to solve all previously learned tasks \cite{nguyen2017}. Formally, we assume that for each task \(\T_{k}\), the data \(\{(\x_{k, i}, y_{k, i})\}_{i = 1}^{n_k}\) are drawn i.i.d. from some distribution \(\D_k\). We denote the model parameters after \(t\) iterations on task \(k\) by \(f_{k}^{(t)}\), and then use \(f^*_{k}\) to refer to the final parameters of the trained model after seeing k tasks.

\subsection{Probabilistic Setup}

When we train a neural network for classification, we implicitly fit a parametric family of probability distributions over the class labels. More formally, a multi-layer perceptron with softmax outputs defines
\begin{equation}
    p\left( y \,\middle|\, \x \mathpunct{;} \w \right)
 = \frac{\exp(f_{y}(\x; \w))}{\sum_{\text{c}} \exp(f_{\text{c}}(\x; \w))}
\end{equation}
where \(f_{\text{c}}(\x; \w)\) is the logit for class \(\text{c}\). Our objective is to maximize the log-likelihood (equivalent to minimizing cross-entropy) so the learning problem lives on the manifold of all these predictive distributions. This is what motivates the use of the Fisher information matrix \cite{amari1998natural}:
\begin{align}
\mathbf{F}(\mathbf{w}) &= \mathbb{E}_{\mathbf{x} \sim \mathcal{D},\; y \sim p\left( y \,\middle|\, \mathbf{x} \mathpunct{;} \mathbf{w} \right)}
\left[ \nabla_{\mathbf{w}} \log p\left( y \,\middle|\, \mathbf{x} \mathpunct{;} \mathbf{w} \right)
 \cdot \nabla_{\mathbf{w}} \log p\left( y \,\middle|\, \mathbf{x} \mathpunct{;} \mathbf{w} \right)^\top \right]
\end{align}
Amari \cite{amari1998natural} demonstrates that a Riemannian structure is given to the parameter space of multilayer networks by the Fisher information matrix. Intuitively, $\mathbf{F}(\mathbf{w})$ measures the sensitivity of the model's output distribution to small changes in each parameter direction around $\mathbf{w}$. As $\mathbf{F}(\mathbf{w})$ is the Hessian of the KL divergence between the distribution parametrized by $\mathbf{w}$ and some fixed distribution, directions of larger Fisher curvature correspond to larger changes in the output distribution.
This geometric structure on our parameter space motivates the natural gradient
\begin{equation}
    \delta(\w) = \F^{-1}(\w)\nabla_{\w} \mathcal{L}(\w)
\end{equation}
as being the steepest-descent direction of loss function \(\mathcal{L}(\w)\) while also constraining the KL-divergence between output distributions. The preconditioning by the inverse Fisher matrix provides us with optimization steps that are invariant to model reparameterization, which can be useful in tasks such as classification \cite{martens2020new}.

\subsection{Orthogonal Gradient Descent}

In continual learning, Orthogonal Gradient Descent (OGD) is a method that aims to prevent forgetting by projecting each new task gradient onto a subspace spanned by previous task gradients, in order to retain performance on previously seen tasks. We begin by restating the original algorithm and then summarize its key theoretical guarantees.

\begin{algorithm}[H]
\caption{Orthogonal Gradient Descent}
\label{alg:ogd}
\begin{algorithmic}[1]
\State \textbf{Input:} Task sequence \(\T_1, \T_2, \T_3, \ldots\), learning rate \(\eta\)
\State \textbf{Output:} The optimal parameter \(\w\)
\State Initialize \(S \gets \{\}\); \(\w \gets \w_0\)
\For{Task ID \(k = 1, 2, 3, \ldots\)}
    \Repeat
        \State \(\g \gets\) Stochastic/Batch Gradient for \(\T_k\) at \(\w\)
        \State \(\tilde{\g} = \g - \sum_{\boldv \in S} \text{proj}_{\boldv}(\g)\)
        \State \(\w \gets \w - \eta \tilde{\g}\)
    \Until{convergence}
    \For{\((\x, y) \in \T_k\) and \(k \in [1, c]\) such that \(y_k = 1\)}
        \State \(\boldu \gets \nabla f_k(\x; \w) - \sum_{\boldv \in S} \text{proj}_{\boldv}\left( \nabla f_k(\x; \w) \right)\)
        \State \(S \gets S \cup \{\boldu\}\)
    \EndFor
\EndFor
\end{algorithmic}
\end{algorithm}

Algorithm \ref{alg:ogd} (shown above) is the original OGD algorithm \cite{OGD}. To justify the use of orthogonal gradient descent, it has been shown that every model update \(\tilde{\g}\) remains a valid descent direction and that the generalization gap can be bounded just as tightly as it is for SGD \cite{bennani}. The forgetting bound is also tightly controlled by a factor that can limit how much any one past task's loss can rise. Together, these results show that OGD is not just empirically effective, but also comes with provable convergence, bounded generalization error, and guaranteed forgetting control \cite{OGD}. All these aspects make it theoretically sound and appealing for continual learning and our applications with the natural gradient.

\section{Orthogonal Natural Gradient Descent}

We now describe our initial attempt to combine the core ideas of OGD and natural gradients to form a new continual learning algorithm  ONG: Orthogonal Natural Gradient Descent. This algorithm aims to combine the parameter space invariant updates of natural gradients with the interference-free projections of orthogonal gradient descent (OGD). The key idea is partitioned into several steps.

\subsection{Natural Preconditioning}
For each new task gradient, \(\g = \nabla_{\w} \mathcal{L}_{k}(\w)\),  we modify this gradient to instead be \(\g = \widehat{\mathbf{F}}(\mathbf{w})^{-1}\ \nabla_{\w} \mathcal{L}_{k}(\w)\)
where $\widehat{\mathbf{F}}(\mathbf{w})$ is an approximation of the Fisher information matrix for the current set of model parameters. This step moves the parameters \(\w\) along the steepest direction on the manifold of the predictive distributions, which minimizes the KL-divergence from the conditional distribution parameterized by the previous model and ensures reparameterization invariance. The remaining steps from the previous algorithm remain the same. We still maintain an orthogonal basis \(S\) that spans the space of previous task gradients and project \(\g\) into the orthogonal complement of span(\(S\)):
\begin{equation}
    \tilde{\g} = \g - \sum_{\boldv \in S} \text{proj}_{\boldv}(\g)
\end{equation}
We perform the same parameter update \(\w \;\leftarrow\; \w - \eta\,\tilde{\g}\),
and, analogous to OGD, once task~\(T_k\) converges, we extract each logit‐wise gradient \(\nabla f_k(\x;\,\w)\), precondition this gradient, project it onto the orthogonal complement of the subspace spanned by all previously added logit‐wise gradients, and denote the result by \( \boldu \). We add \(\boldu\) to our set \(S\), maintaining \(S\) as an orthogonal basis of “sensitive directions.” Thus, the set \(S\) thus spans the subspace of all previously seen \emph{natural} gradients.


\subsection{OGD-Plus (OGD+)}
OGD-Plus (OGD+) is an algorithm that enhances vanilla OGD by storing the feature maps with respect to samples from previous tasks, as well as the feature maps with respect to samples from the current task \cite{bennani}. These extra samples are saved in a dedicated memory, the sample memory. The authors motivated this change in order to make OGD more robust to the NTK (Neural Tangent Kernel) variation phenomena in their experiments. The changes from OGD to OGD+ can be found in blue in the below algorithm.

\begin{algorithm}[H]
\caption{ONG / ONG+}
\label{alg:ong}
\begin{algorithmic}[1]
\State \textbf{Input:} Task sequence \(\T_1, \T_2, \T_3, \ldots\), learning rate \(\eta\)
\State \textbf{Output:} The optimal parameter \(\w\)
\State Initialize \(S \gets \{\}\); \textcolor{blue}{\(S_D \gets \{\}\)}; \(\w \gets \w_0\)
\For{Task ID \(k = 1, 2, 3, \ldots\)}
    \Repeat
        \State \(\g \gets\) Stochastic/Batch Gradient for \(\T_k\) at \(\w\)
        \State \textcolor{red}{\(\g \gets\F^{-1} \g\)}
        \State \(\tilde{\g} = \g - \sum_{\boldv \in S} \text{proj}_{\boldv}(\g)\)
        \State \(\w \gets \w - \eta \tilde{\g}\)
    \Until{convergence}
    \State \textcolor{blue}{Sample \(H \subset S_D\)}
    \For{ \((\x, y) \in \T_k \textcolor{blue}{\cup}\ \textcolor{blue}{H}\)
 and \(k \in [1, c]\) such that \(y_k = 1\)}
        \State \textcolor{red}{\(\g \gets\F^{-1} \nabla f_k(\x; \w) \)}
        \State  \( \boldu \gets \textcolor{red}{\g} - \sum_{\boldv \in S}
        \mathrm{proj}_{\boldv}\left( \textcolor{red}{\g} \right)  \)
        \State \(S \gets S \cup \{\boldu\}\)
    \EndFor
    \State \textcolor{blue}{Sample \(D \subset \T_k\)}
    \State \textcolor{blue}{Update \(S_D \gets S_D \cup D\)}
\EndFor
\end{algorithmic}
\end{algorithm}



\subsection{ONG and ONG+}
In this paper, we present ONG and ONG+, extensions of OGD and OGD+ using natural gradients, respectively. Our changes are in red in Algorithm \ref{alg:ong}: we precondition gradients using the Fisher information matrix both when updating model parameters and when storing task gradients in memory (the span of this set is what future gradients are ultimately projected onto).

\subsection{Theoretical Guarantees}
This section focuses on studying the ONG algorithm (aka Algorithm 2 without the blue text) to obtain some theoretical results.
Analogous to Lemma 3.1 from the original OGD paper \cite{OGD}, we present a proof that the preconditioned, projection-subtracted gradient \(\tilde{\g}\) is still a descent direction with respect to the Fisher metric. 

As described in section 3.1, in the context of probabilistic models, the Riemannian metric tensor is the Fisher information matrix \(\F\). Theorem 1 from \cite{OGD} highlights that in this Riemannian manifold, \(-\F^{-1}\nabla \mathcal{L}(\w)\) is the steepest descent direction of the loss. Thus, if \(\langle \tilde{\g}, -\F^{-1}\nabla_{\w} \mathcal{L}(\w) \rangle \leq 0, \tilde{\g}\) will also be a valid descent direction with respect to the metric structure of the manifold. The following lemma proves this.

\begin{lemma}
\label{lem:descent}

Let \(\F^{-1} \g\) be the natural gradient of loss function \(\mathcal{L}(\w)\) and \(S=\{\boldv_1, \ldots, \boldv_n\}\) be an orthogonal basis. Define \(\tilde{\g} = \F^{-1} \g - \sum_i^k {\mathrm{proj}_{\boldv_i}(\F^{-1} \g)}\), where \({\mathrm{proj}_{\boldv_i}(\F^{-1} \g)}\) denotes the projection of \(\F^{-1}\g\) onto the \(\boldv_i\). Then, \(-\tilde{\g}\) is also a descent direction for \(\mathcal{L}(\w)\), with respect to the metric structure of the model parameter space.
\end{lemma}

The proof for this lemma can be found in Appendix \ref{sec:lemma4.1pf}.




\subsection{EKFAC Fisher Approximation}
Calculating the exact Fisher information matrix (FIM) and its inverse becomes prohibitively expensive as our model size grows. To make computation tractable, we utilize the Eigenvalue-corrected Kronecker Factorization (EKFAC) \cite{George2018EKFAC}. EKFAC works by providing a diagonal approximation to the FIM not in the raw parameter basis, but in the Kronecker-factored eigenbasis, where a diagonal captures most of the curvature information at a very low cost. Each neural net layer's Fisher block is approximated as \(\F = \boldsymbol{A} \otimes \boldsymbol{B}\) with the Kronecker factors \(\boldsymbol{A} = \boldsymbol{Q}_{\boldsymbol{A}}\boldsymbol{\Lambda_A}\boldsymbol{Q_A^\top} \text{ and } \boldsymbol{B} = \boldsymbol{Q_B}\boldsymbol{\Lambda_B}\boldsymbol{Q_B^\top}\). We track just the diagonal matrices \(\boldsymbol{\Lambda_A}\) and \(\boldsymbol{\Lambda_B}\) in these eigenbases, so each mini-batch update just becomes two small eigenbasis transformations plus a diagonal rescaling. We utilize this approximation of the Fisher matrix in our code implementation. We adapt this EKFAC algorithm \cite{wiseodd2020naturalgradients} in our code implementation for approximating the Fisher matrix, leading to faster and more stable training.

\section{Experiments}
\label{sec:experiments}

\subsection{Datasets}
Our preliminary evaluations include two standard benchmarks utilized extensively in continual learning literature: Permuted MNIST and Rotated MNIST. Permuted MNIST generates \(K\) tasks by fixing \(K\) independent random permutations of the \(784\) input pixels. Each task is then the original MNIST images with its pixels reshuffled by one of those permutations, forcing the model to learn to classify under radically different orderings without forgetting prior permutations. 
Rotated MNIST applies a fixed rotation angle \(\theta_k\) to every MNIST digit. At task \(k\), all training and test images are rotated by \(\theta_k\), which forces the model to continually adapt to new orientations while remembering earlier orientations with different angles. 

\subsection{Evaluation Criteria}
For evaluation purposes, we adapt the same metrics as \cite{bennani}: average accuracy and average forgetting. Let \(a_{T, k}\) denote the accuracy of the model on the \(\T_k\) after being trained on the task \(\T_T\). Average accuracy (\(A_T\)) is the accuracy of the model after its trained on task \(\T_T\). It is defined as:
\begin{equation}
    A_T = \frac{1}{T}\sum_{k = 1}^{T} a_{T, k}
\end{equation} Average forgetting (\(F_T\)) is the average forgetting the model has after its been trained on task \(\T_T\). It is defined as:
\begin{equation}
    F_T = \frac{1}{T - 1}\sum_{k = 1}^{T - 1} \max_{t \in \{1, ... T - 1\}} (a_{t, k} - a_{T, k})
\end{equation}

\subsection{Setup}
Details can be found in Appendix \ref{sec:setup}.

\subsection{Results}
\subsubsection{Forgetting and Average Validation Accuracy}
Here, we present the average forgetting and average validation accuracy, as described above, of the two baseline methods (OGD and OGD+), as well as our natural gradient extensions to each (ONG and ONG+). Table \ref{tab:exp:bench-Forgetting} highlights the average forgetting values and Table \ref{tab:exp:bench-Accuracy} highlights the final average validation accuracy. As we can see, simply adding a preconditioner to the OGD algorithm leads to extremely high fogetting and lower accuracy. These results suggest that there is some underlying tension between the Euclidian-style projections of OGD with the Fisher-geometry descent directions. While the reason for this discrepancy is unclear, we are currently investigating this phenomena by using crafted datasets and evaluations.

\begin{table}[h]
\centering
\caption{Average forgetting of all the methods we considered (lower is better).}
\label{tab:exp:bench-Forgetting}
\begin{tabular}{lllll}
\toprule
& \textbf{Permuted MNIST} &   \textbf{Rotated MNIST} & \\
\midrule

\textbf{OGD} &  4.1534 &  16.1282 \\
\midrule

\textbf{OGD+} &  1.0563 &  6.0064 \\
\midrule
\textbf{ONG} &  58.7240 &  28.7360 \\
\midrule
\textbf{ONG+} &  57.5692 &  27.6238 \\
\midrule
\bottomrule
\end{tabular}
\end{table}
\vspace{-1em}
\begin{table}[h]
\centering
\caption{Final validation accuracy, averaged over all tasks, of all the methods we considered.}
\label{tab:exp:bench-Accuracy}
\begin{tabular}{lllll}
\toprule
 & \textbf{Permuted MNIST} &   \textbf{Rotated MNIST} & \\
\midrule

\textbf{OGD} &  82.6604 &  77.7443 \\
\midrule

\textbf{OGD+} &  85.8746 &  87.6950 \\
\midrule
\textbf{ONG} &  32.5053 &  69.0968 \\
\midrule
\textbf{ONG+} &  33.5723 &  70.1771 \\
\midrule
\bottomrule
\end{tabular}
\end{table}

\subsubsection{Permuted MNIST}
\begin{figure}[htbp]
  \centering
  \begin{subfigure}[t]{0.48\linewidth}
    \centering
    \includegraphics[width=\linewidth,valign=t]{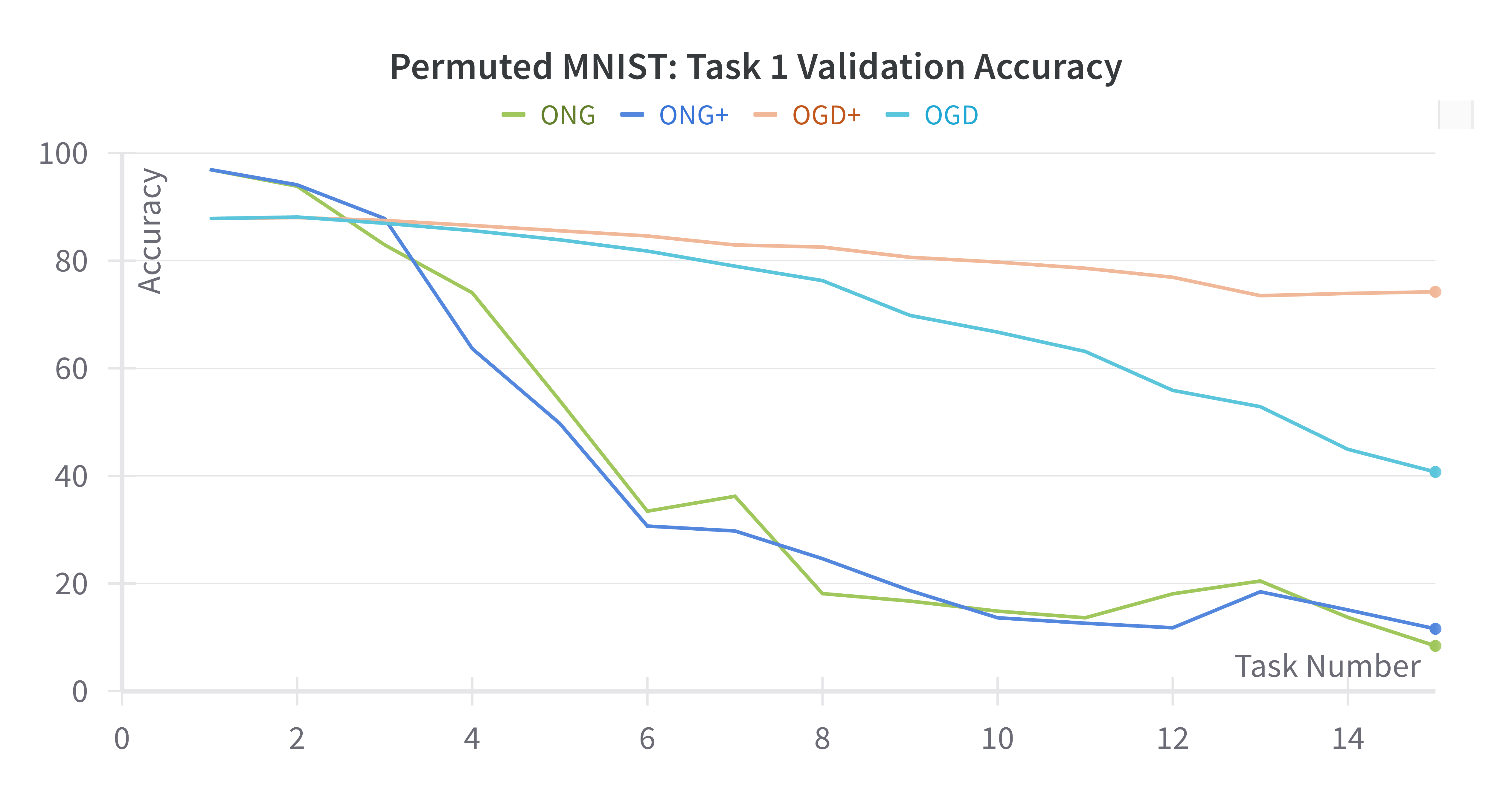}
    \caption{Task 1 accuracy throughout tasks}
    \label{fig:acc-task1}
  \end{subfigure}
  \hfill
  \begin{subfigure}[t]{0.48\linewidth}
    \centering
    \includegraphics[width=\linewidth,valign=t]{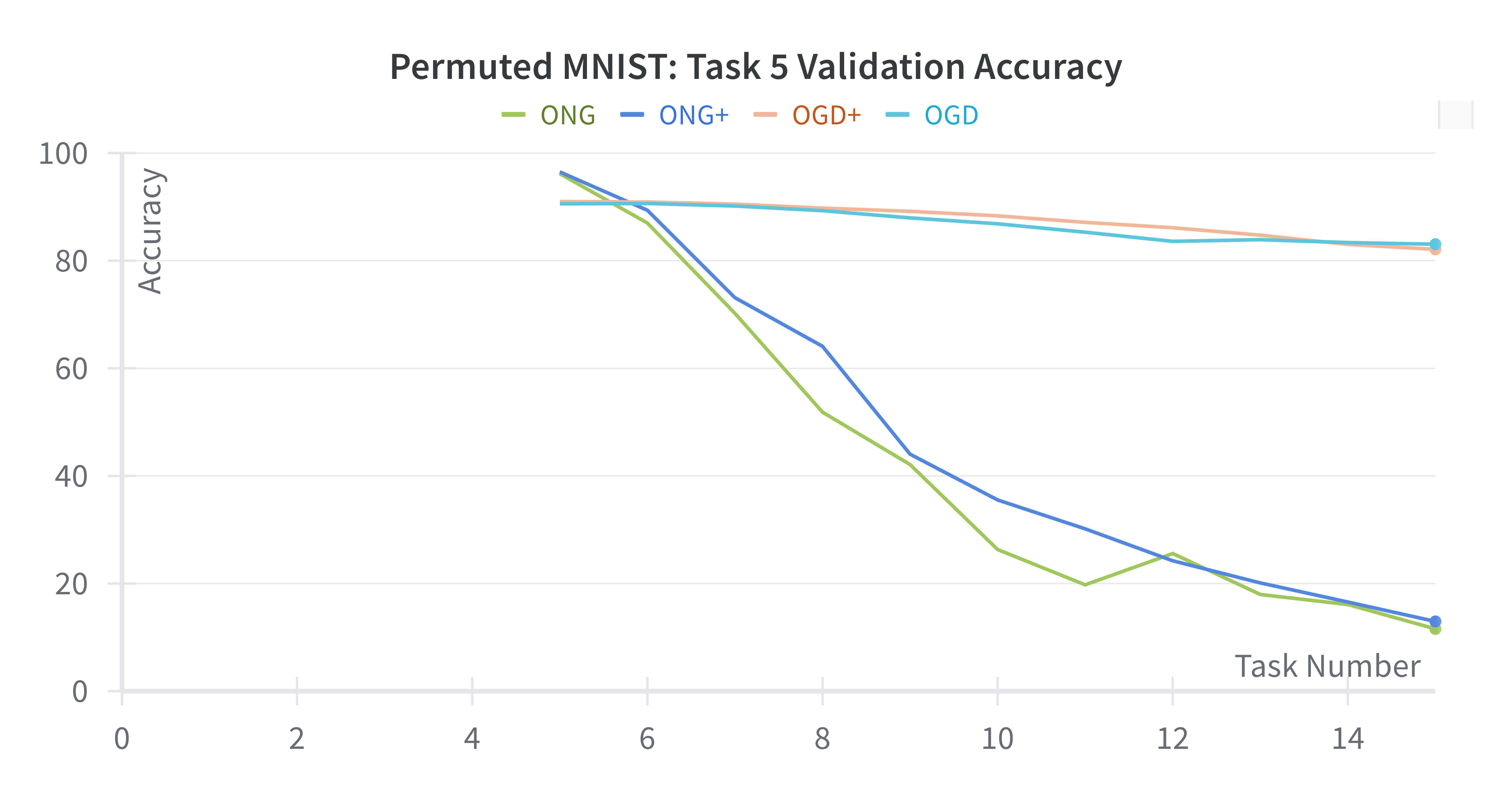}
    \caption{Task 5 accuracy throughout tasks}
    \label{fig:acc-task2}
  \end{subfigure}

  \vspace{1em}

  \begin{subfigure}[t]{0.48\linewidth}
    \centering
    \includegraphics[width=\linewidth,valign=t]{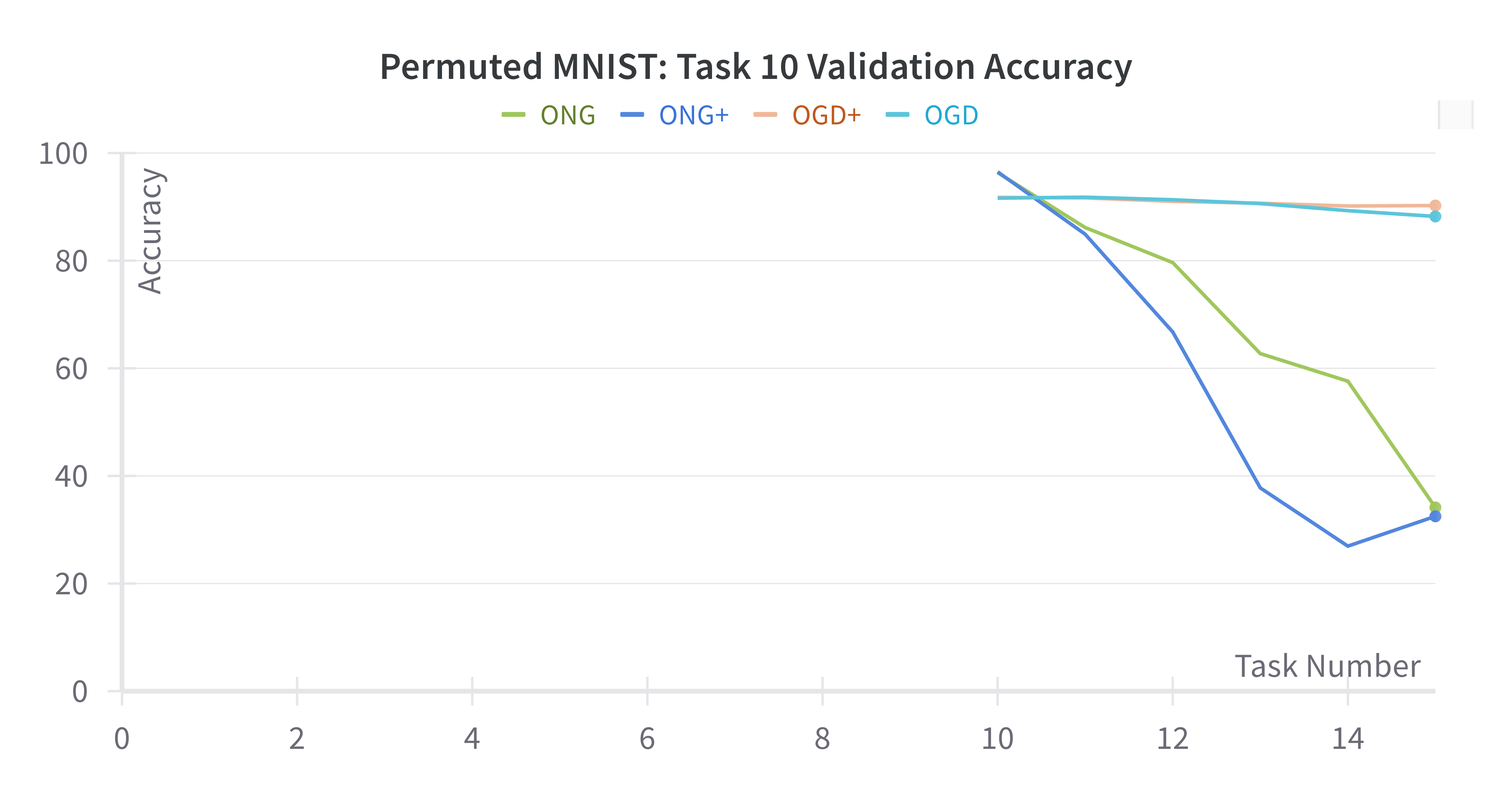}
    \caption{Task 10 accuracy throughout tasks}
    \label{fig:acc-task3}
  \end{subfigure}
  \hfill
  \caption{Validation Accuracy for tasks 1, 5, and 10 of the Permuted MNIST dataset throughout training. The model is sequentially trained on tasks 1 through 15, and thus the x-axis is presented in terms of number of tasks.}
  \label{fig:all-accuracies-permuted}
\end{figure}

Here, we more closely examine our methods' performance on Permuted MNIST. Table \ref{tab:app-permuted-mnist} lists the final task-wise test accuracies for each method after the model finishes sequentially training on all tasks.  The plots in Figure \ref{fig:all-accuracies-permuted} examine the evolution of validation accuracy during training, specifically how per-task accuracies evolve as the model continues to be trained on newer and newer tasks. These finer-grain results show just how quickly the accuracy starts dropping off as we train on newer tasks, and further reinforce the idea that our method of projection is likely misaligned with our metric space of choice.
\begin{table}[H]
  \centering
  \caption{Permuted MNIST: Test accuracy of each method on the indicated task after training sequentially on all tasks.}
  \label{tab:app-permuted-mnist}

  \begin{subtable}[t]{\linewidth}
    \centering
    \caption{Tasks 1\,--\,7}
    \label{tab:perm-mnist-1-7}
    \begin{tabular}{l *{7}{c}}
      \toprule
      Method     & \multicolumn{7}{c}{Accuracy} \\
      \cmidrule(lr){2-8}
                & Task 1 & Task 2 & Task 3 & Task 4 & Task 5 & Task 6 & Task 7 \\
      \midrule
      OGD    & 38.91 & 68.55 & 71.73 & 79.71 & 83.41 & 87.27 & 85.83 \\
      OGD+   & 74.69 & 76.57 & 76.12 & 79.42 & 82.03 & 84.72 & 85.16 \\
      ONG    &  5.44 & 10.22 & 17.08 & 14.24 &  8.17 & 12.88 & 15.74 \\
      ONG+   & 13.64 & 16.48 & 15.44 & 16.26 & 11.00 & 14.39 & 19.69 \\
      \bottomrule
    \end{tabular}
  \end{subtable}

  \vspace{1em}

  \begin{subtable}[t]{\linewidth}
    \centering
    \caption{Tasks 8\,--\,15}
    \label{tab:perm-mnist-8-15}
    \begin{tabular}{l *{8}{c}}
      \toprule
      Method       & \multicolumn{8}{c}{Accuracy} \\
      \cmidrule(lr){2-9}
                & Task 8 & Task 9 & Task 10 & Task 11 & Task 12 & Task 13 & Task 14 & Task 15 \\
      \midrule
      OGD    & 87.26 & 88.79 & 88.59 & 89.93 & 91.56 & 92.02 & 92.96 & 93.39 \\
      OGD+   & 87.91 & 89.03 & 90.56 & 90.74 & 91.92 & 92.74 & 93.11 & 93.39 \\
      ONG    & 23.73 & 24.75 & 25.89 & 44.99 & 40.09 & 66.53 & 81.48 & 96.35 \\
      ONG+   & 23.86 & 26.40 & 33.47 & 29.67 & 43.95 & 54.85 & 88.15 & 96.34 \\
      \bottomrule
    \end{tabular}
  \end{subtable}
\end{table}

\subsubsection{Rotated MNIST}
\begin{figure}[htbp]
  \centering
  \begin{subfigure}[t]{0.48\linewidth}
    \centering
    \includegraphics[width=\linewidth,valign=t]{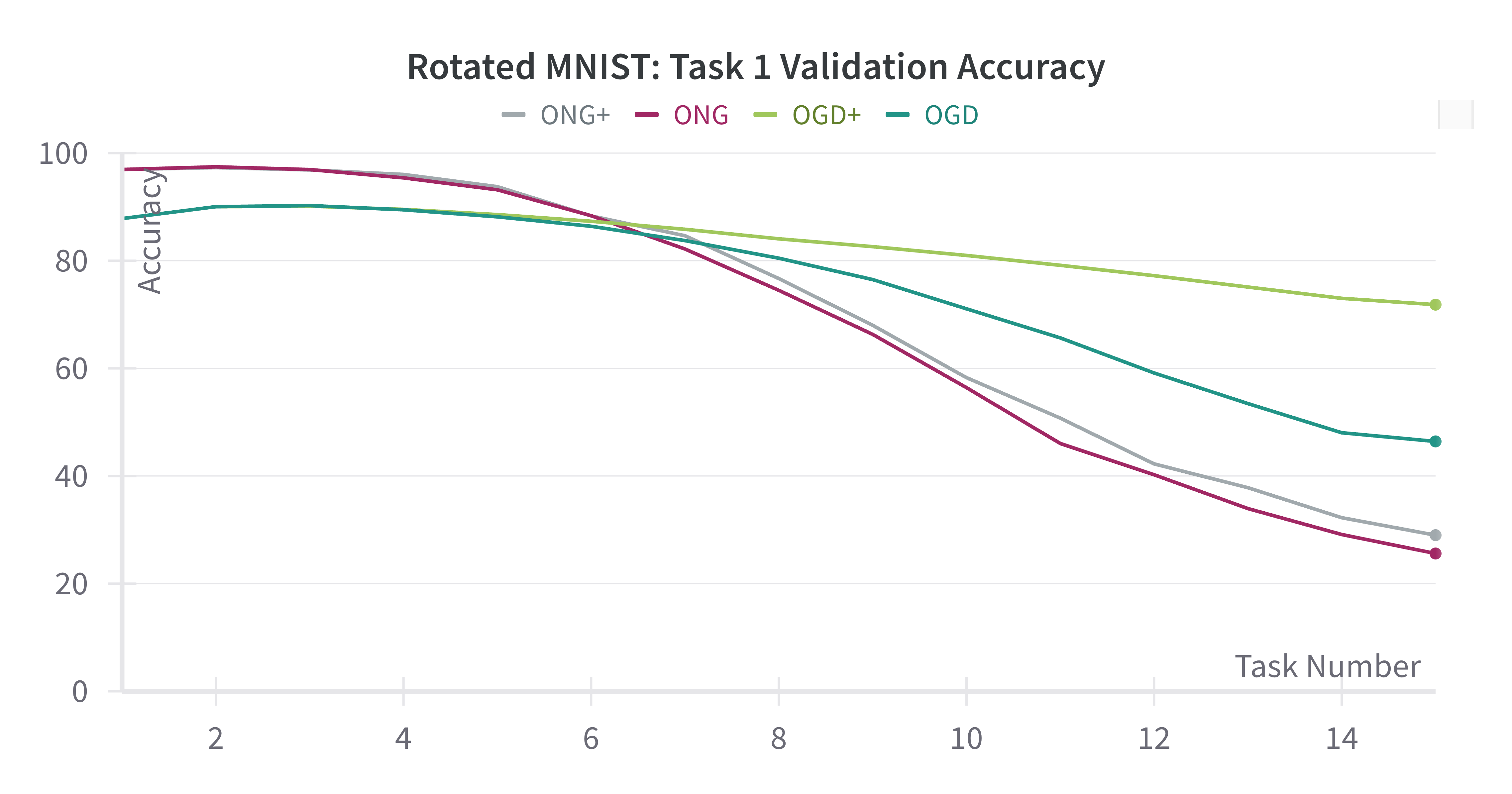}
    \caption{Task 1 accuracy throughout tasks}
    \label{fig:acc-task1}
  \end{subfigure}
  \hfill
  \begin{subfigure}[t]{0.48\linewidth}
    \centering
    \includegraphics[width=\linewidth,valign=t]{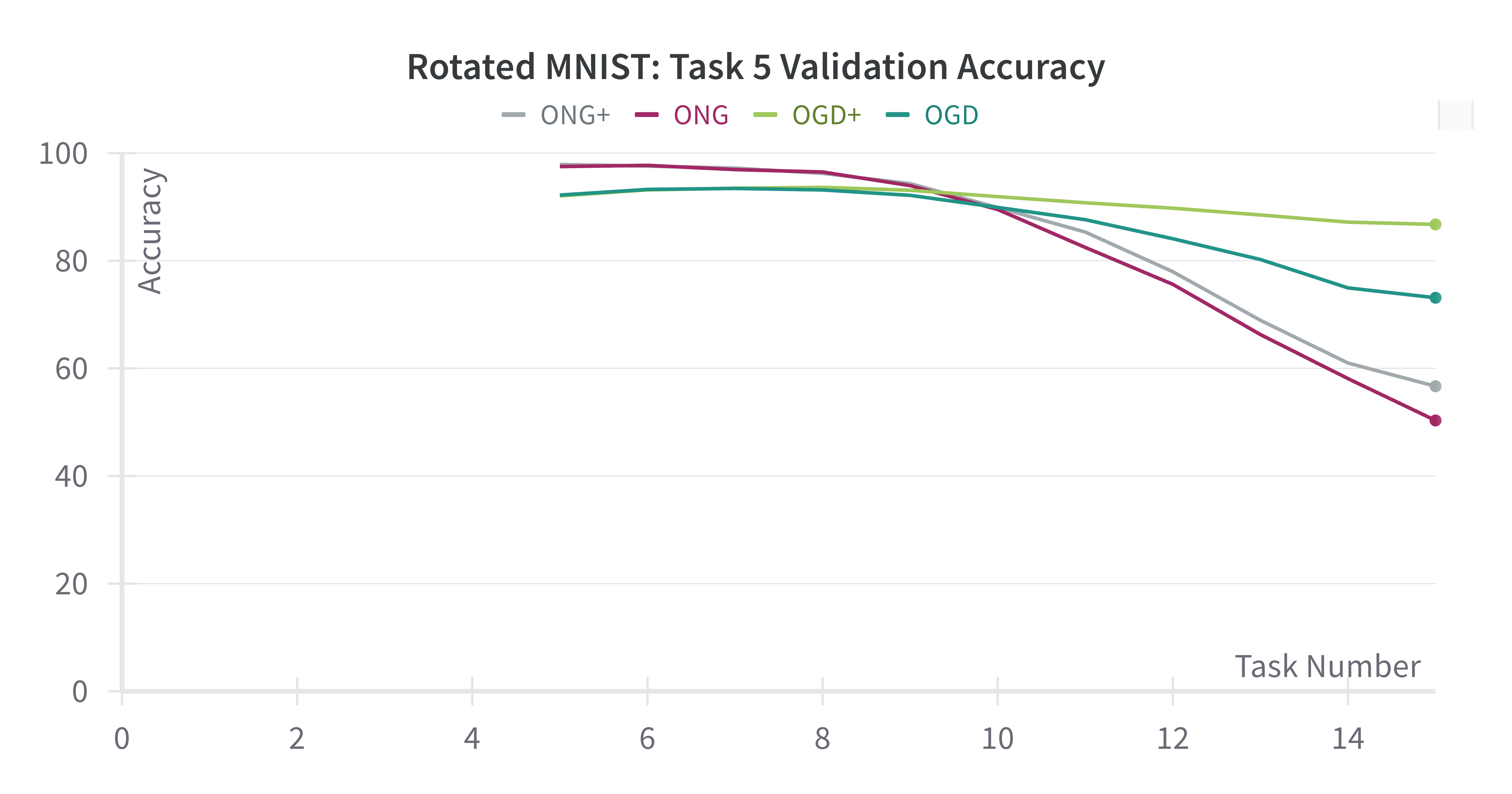}
    \caption{Task 5 accuracy throughout tasks}
    \label{fig:acc-task2}
  \end{subfigure}

  \vspace{1em}

  \begin{subfigure}[t]{0.48\linewidth}
    \centering
    \includegraphics[width=\linewidth,valign=t]{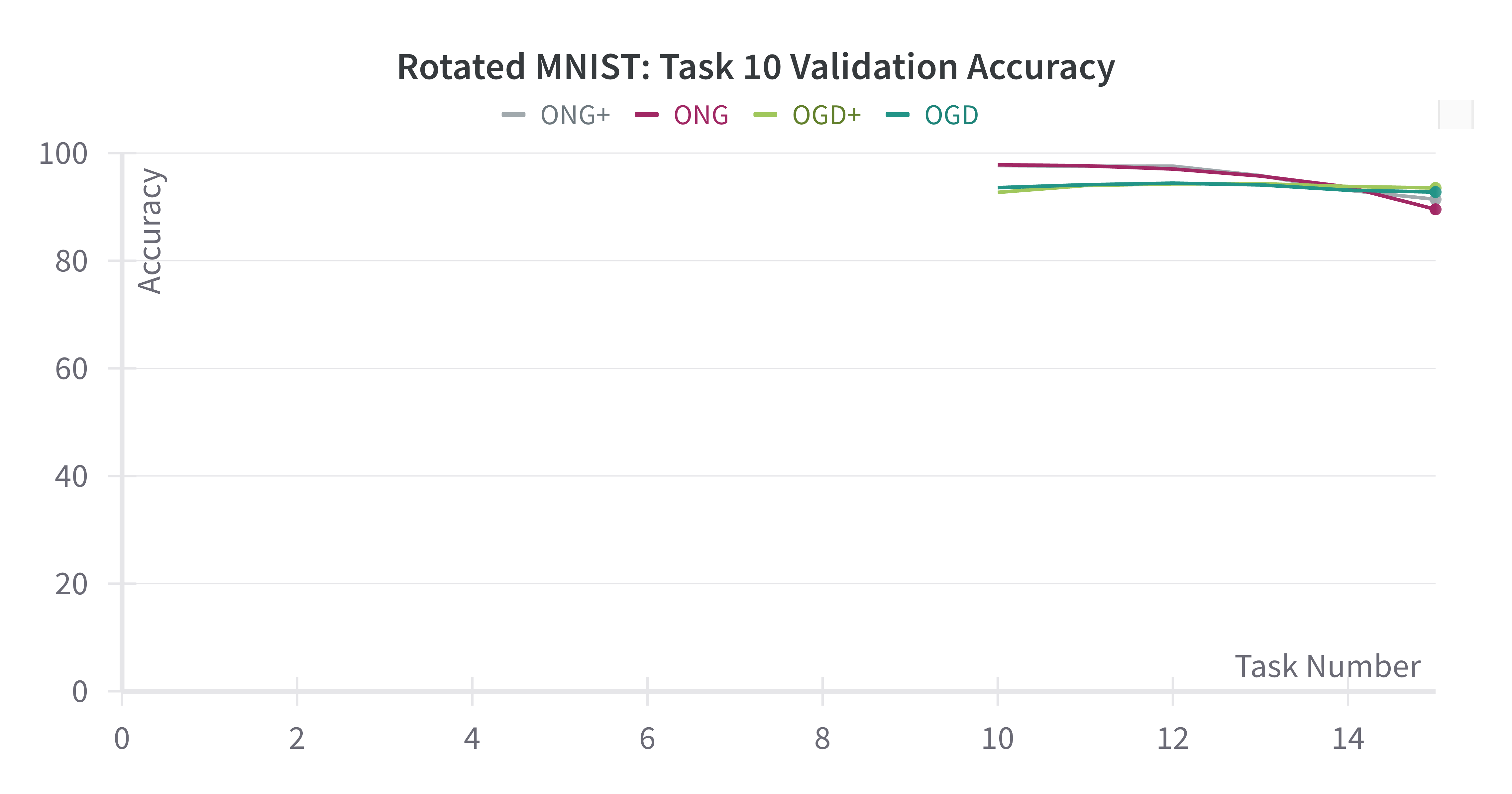}
    \caption{Task 10 accuracy throughout tasks}
    \label{fig:acc-task3}
  \end{subfigure}
  \hfill
  \caption{Validation Accuracy for tasks 1, 5, and 10 of the Rotated MNIST dataset throughout training. The model is sequentially trained on tasks 1 through 15, and thus the x-axis is presented in terms of number of tasks.}
  \label{fig:all-accuracies-rotated}
\end{figure}

Here, we more closely examine our methods' performance on Rotated MNIST. Table \ref{tab:app-rotated-mnist} lists the final task-wise test accuracies for each method after the model finishes sequentially training on all tasks. The plots in Figure \ref{fig:all-accuracies-rotated} examine the evolution of validation accuracy during training, specifically how per-task accuracies evolve as the model continues to be trained on newer and newer tasks. While less extreme, we still observe the same sharper dropoff here as well.
\begin{table}[H]
  \centering
  \caption{Rotated MNIST: Test accuracy of each model on the indicated task after training sequentially on all tasks.}
  \label{tab:app-rotated-mnist}

  \begin{subtable}[t]{\linewidth}
    \centering
    \caption{Tasks 1\,--\,7}
    \label{tab:rot-mnist-1-7}
    \begin{tabular}{l *{7}{c}}
      \toprule
      Method     & \multicolumn{7}{c}{Accuracy} \\
      \cmidrule(lr){2-8}
                & Task 1 & Task 2 & Task 3 & Task 4 & Task 5 & Task 6 & Task 7 \\
      \midrule
      OGD    & 41.83 & 43.75 & 53.89 & 61.19 & 68.45 & 74.93 & 82.01 \\
      OGD+   & 69.67 & 73.21 & 79.06 & 83.27 & 85.71 & 86.87 & 89.59 \\
      ONG    & 24.42 & 27.60 & 34.24 & 41.48 & 49.81 & 58.32 & 67.85 \\
      ONG+   & 27.59 & 30.30 & 37.41 & 43.31 & 51.86 & 59.68 & 68.94 \\
      \bottomrule
    \end{tabular}
  \end{subtable}

  \vspace{1em}

  \begin{subtable}[t]{\linewidth}
    \centering
    \caption{Tasks 8\,--\,15}
    \label{tab:rot-mnist-8-15}
    \begin{tabular}{l *{8}{c}}
      \toprule
      Method     & \multicolumn{8}{c}{Accuracy} \\
      \cmidrule(lr){2-9}
                & Task 8 & Task 9 & Task 10 & Task 11 & Task 12 & Task 13 & Task 14 & Task 15 \\
      \midrule
      OGD    & 85.49 & 89.33 & 91.97 & 93.39 & 94.83 & 95.03 & 95.27 & 94.80 \\
      OGD+   & 90.15 & 91.91 & 92.95 & 93.74 & 94.83 & 95.15 & 94.92 & 94.40 \\
      ONG    & 75.81 & 84.03 & 90.02 & 93.88 & 96.41 & 97.17 & 97.74 & 97.68 \\
      ONG+   & 77.10 & 84.27 & 89.58 & 93.64 & 96.23 & 97.39 & 97.58 & 97.78 \\
      \bottomrule
    \end{tabular}
  \end{subtable}
\end{table}

\newpage
\section{Conclusion and Outlook}    
In this work, we proposed and investigated Orthogonal Natural Gradient Descent (ONG), a novel algorithm that incorporates natural gradients with OGD, a continual learning algorithm.

Our theoretical contributions include proving that ONG maintains descent-direction guarantees under the Fisher metric and detailing an efficient implementation using the EKFAC approximation.

Our empirical evaluation, however, yielded a counterintuitive yet insightful result: ONG underperforms its Euclidean counterpart on standard continual learning benchmarks. Rather than a simple failure of the method, we interpret this as a key finding of our study: \textbf{there exists a fundamental tension between Fisher preconditioning and standard orthogonal projections}. This discovery suggests that a naive combination of these methods is not only suboptimal but can be detrimental, indicating that the underlying geometry is not immediately compatible. This finding motivates a clear path for future research focused on reconciling these two perspectives. Our primary hypothesis is that the type of projection we utilize must be chosen to be compatible with the descent directions natural gradient descent uses. We are currently investigating on understanding what the exact failure modes are by crafting synthetic, task-datasets and analyzing the training dynamics. Further future work will involve looking more closely at the evolution of the task-specific gradient subspaces, and the accuracy of the Fisher approximation estimates. Ultimately, we aim to utilize these insights and develop a geometrically-meaningful way of performing "projection" with natural gradients that preserves previously seen gradient directions and enables high accuracy on older tasks.

To that end, incorporating more advanced concepts from Manifold Learning and Information Geometry presents an exciting direction. The concept of parallel transport (see Appendix \ref{sec:plleltspt} for its definition) is particularly promising, as it provides a principled way to move tangent vectors (i.e., gradients) between different points on the parameter manifold \cite{budninskiy2018parallel, li2022federated}. By using parallel transport to bring gradients from previous tasks into the same tangent space as the current gradient, it may be possible to define a geometrically consistent inner product and basis for projection.

Additional avenues for research include a more rigorous study of the parameterization-invariance property, a primary motivator for this work, to understand which kind of model reparameterizations our method is robust to. Finally, we posit that the limitations we observed may be partially attributable to the high task correlation in benchmarks like Permuted and Rotated MNIST. We hypothesize that after developing a more roboust, geometrically-aware continual learning method, its benefits will be more pronounced in more challenging and realistic continual learning scenarios involving diverse, uncorrelated task sequences. Future work will extend our empirical validation to these more complex domains.

\newpage
\bibliographystyle{unsrtnat}
\bibliography{references}


\newpage

\appendix

\section{Proof of Lemma 4.1}
\label{sec:lemma4.1pf}

For \(- \tilde{\g}\) to be a valid descent direction it should satisfy \(\langle -\tilde{\g}, \F^{-1}\g \rangle \le 0\). We have
\begin{align}
\langle -\tilde{\g},\,\, \F^{-1}\g \rangle & = \langle -\tilde{\g}, \tilde{\g}+\sum_{i=1}^k {\mathrm{proj}_{\boldv_i}(\F^{-1}\g)}\rangle\\
&=-\|\tilde{\g}\|^2-\langle \tilde{\g}, \sum_{i=1}^k {\mathrm{proj}_{\boldv_i}(\F^{-1}\g)}\rangle~.\label{eq:lemma_a2_1}
\end{align}
Since \(\tilde{\g}=\F^{-1}\g-\sum_{i=1}^k {\mathrm{proj}_{\boldv_i}(\F^{-1}\g)}\) is orthogonal to the space spanned by the vectors in \(S\) and each \(\mathrm{proj}_{\boldv_i}(\F^{-1}\g)\) lies in this space, \(\tilde{\g}\) will be orthogonal to \(\sum_{i=1}^k {\mathrm{proj}_{\boldv_i}(\F^{-1}\g)}\). Thus, \(
    \langle \tilde{\g}, \sum_{i=1}^k {\mathrm{proj}_{\boldv_i}(\F^{-1}\g)}\rangle=0\).
Substituting this into Eq.~\ref{eq:lemma_a2_1}, we arrive at \(
    \langle -\tilde{\g},\,\, \F^{-1}\g \rangle=-\|\tilde{\g}\|^2\le0\).
Therefore, \(-\tilde{\g}\) is a valid descent direction for \(\mathcal{L}(\w)\) while maintaining orthogonality to \(S\). \(\blacksquare\)

\section{Setup}
\label{sec:setup}

We use the official code implementation of \cite{bennani}, which can be obtained \href{https://github.com/MehdiAbbanaBennani/continual-learning-ogdplus?tab=readme-ov-file}{here} as our starting point for building ONG. We also adapt the EKFAC algorithm provided by \cite{wiseodd2020naturalgradients} for our efficient Fisher information matrix implementation. Both repositories have a MIT license.

We train a randomly-initialized MLP (using Kaiming/He initialization) with depth \(L = 3\), widths \(n_0 = 784\), \(n_1 = n_2 = d = 100\), \(n_3 = 10\), and ReLU activation functions. The total number of parameters is \(\sim115,000\).

After some experimentation, we chose 3 epochs per task, a learning rate of \(0.001\), a memory buffer size of \(100\) samples per task, and a batch size \(B = 32\). No training augmentations were added (though this could be part of a future follow-up). All training and validation was done on a single NVIDIA A100.

The training step logic was modified to keep track of the running estimates needed for the EKFAC algorithm. The optimizer step logic was modified to efficiently use EKFAC for preconditioning the gradient vector.

\section{Parallel Transport}
\label{sec:plleltspt}
\begin{definition}[Parallel transport]
  Given a complete Riemannian manifold \((\mathcal{M},g)\) and two points \(x,y\in\mathcal{M}\), the \emph{parallel transport}
  \begin{equation}
    P_{x\to y} : T_{x}\mathcal{M} \;\longrightarrow\; T_{y}\mathcal{M}
    \footnote{On a complete Riemannian manifold the geodesic between any two points is unique, so \(P_{x\to y}\) is well-defined.}
  \end{equation}
  is the linear isometry satisfying
  \begin{equation}
    \bigl\langle P_{x\to y}\,\xi,\;P_{x\to y}\,\zeta\bigr\rangle_{y}
    \;=\;
    \langle \xi,\zeta\rangle_{x}
    \quad
    \forall\,\xi,\zeta\in T_{x}\mathcal{M}.
  \end{equation}
\end{definition}

\newpage

\end{document}